%% file: main.tex
\documentclass[twocolumn,twoside]{Jornadas}
\usepackage[utf8]{inputenc}
\usepackage[spanish]{babel}
\usepackage{listings}
\usepackage{algorithm}
\usepackage{amsmath}
\usepackage{algorithmicx}
\usepackage{algcompatible}
\usepackage{adjustbox}
\usepackage{booktabs} 
\usepackage{graphicx}
\usepackage{color}
\usepackage{caption}
\usepackage{algcompatible}
\usepackage[caption=false,font=footnotesize]{subfig}
\usepackage{placeins}
\usepackage{hyperref}
\usepackage{url}

\definecolor{gray97}{gray}{.97}
\definecolor{gray75}{gray}{.75}
\definecolor{gray45}{gray}{.45}

\def\BibTeX{{\rm B\kern-.05em{\sc i\kern-.025em b}\kern-.08em
    T\kern-.1667em\lower.7ex\hbox{E}\kern-.125emX}}

\graphicspath{{.}{./Figuras/}}

\begin{document}

\title{Estudio de la eficiencia en la escalabilidad de GPU`s para el entrenamiento de Inteligencia Artificial}

\author{%
     David Cortes, Carlos Juiz, Belen Bermejo%
     \thanks{Dpto. de Ciencias Matemáticas e Informática, Universidad de las Islas Baleares, 
     e-mail: {\tt \{d.cortes, cjuiz, bbermejo\}@uib.es}}
}

\maketitle
\markboth{}{}
\pagestyle{empty} 
\thispagestyle{empty} 

\begin{abstract}
El entrenamiento de modelos de aprendizaje profundo a gran escala se ha convertido en un desafío clave para la comunidad científica y la industria. Si bien el uso masivo de GPUs puede acelerar considerablemente los tiempos de entrenamiento, este enfoque conlleva un impacto negativo en la eficiencia. En este artículo, presentamos un análisis detallado de los tiempos presentados por MLPerf Training v4.1 en cuatro cargas de trabajo: BERT, Llama2 LoRA, RetinaNet y Stable Diffusion, evidenciando que existen configuraciones que optimizan la relación entre el rendimiento, uso de GPUs y eficiencia. Los resultados señalan un “punto de equilibrio” que permite disminuir tiempos de entrenamiento manteniendo maximizando la eficiencia.
\end{abstract}

\begin{keywords}
GPUs; Rendimiento; MLPerf; Deep Learning; Benchmarking; Eficiencia; Aceleración 
\end{keywords}


\input{01-Intro.tex}
\input{02-Background.tex}

\input{03-Trabajos-Relacionados.tex}

\input{04-Method.tex}

\input{05-Experimental.tex}
\input{06-Discusion}
\input{07-Conclusions}


\bibliographystyle{Jornadas}
\bibliography{biblio}

\end{document}

%% file: 01-Intro.tex
\section{Introducción}
\PARstart{E}{l}entrenamiento de modelos de aprendizaje profundo de gran escala se ha convertido en una prioridad estratégica tanto en entornos académicos como industriales \cite{Kurgalin2019}. Con la creciente complejidad de tareas como la comprensión del lenguaje natural, la detección de objetos y la generación de contenido, el uso de infraestructuras con un gran número de GPUs se ha consolidado como la vía para reducir significativamente el tiempo de entrenamiento. Sin embargo, la ampliación masiva de recursos no siempre se traduce en una mejora proporcional de la eficiencia, puesto que la sobrecarga de comunicación y la sincronización entre nodos suelen afectar negativamente el rendimiento de cada acelerador.

En este contexto, MLPerf Training \cite{MLSYS2020_411e39b1} ofrece un marco de facto de evaluación estandarizada para comparar diversas configuraciones y arquitecturas de hardware. A través de cargas de trabajo representativas, como BERT, Llama2 LoRA, Retinanet y Stable Diffusion, diferentes empresas han reportado el tiempo que tardan sus máquinas en entrenar estos modelos aplicando el benchmark en sus sistemas. Estos tiempos resultantes se han reportado en un repositorio de acceso libre \cite{LinkMLCommons}. Este estudio presenta un análisis de estos resultados, que evidencian dos tendencias contrapuestas: la disminución del tiempo total de entrenamiento al aumentar el número de GPUs y la consiguiente reducción de la eficiencia por acelerador. En este artículo se aborda el desafío de conciliar estas dos realidades, sugiriendo la existencia de configuraciones equilibradas que combinan tiempos de entrenamiento competitivos con un aprovechamiento óptimo de los recursos, el cual puede repercutir posiblemente en minimizar los costes operativos y el impacto energético. Este enfoque cobra importancia no solo para centros de investigación y laboratorios con grandes presupuestos, sino también para aquellas organizaciones que requieren maximizar su sostenibilidad.

Para abordar este objetivo, en este documento se exponen resultados cuantitativos que ilustran cómo la eficiencia varía en función del número de aceleradores, y se discuten los factores que determinan el rendimiento en diferentes arquitecturas de hardware. Con ello, se pretende contribuir a la toma de decisiones informadas al momento de planificar y dimensionar infraestructuras de cómputo para el entrenamiento de modelos de Deep Learning.

Este artículo se estructura de la siguiente manera: en la Sección \ref{sec2}, se presentan los conceptos fundamentales y métricas clave de rendimiento y escalabilidad; la Sección \ref{sec3} revisa los trabajos relacionados más relevantes; en la Sección \ref{sec4}, se describe la metodología empleada para analizar y comparar los sistemas; la Sección \ref{sec5}, expone los resultados experimentales; en la Sección \ref{sec6}, se amplia la discusión; finalmente en \ref{sec7} se ofrecen las conclusiones y posibles líneas de investigación futura. 

%% file: 02-Background.tex
\section{Conceptos Previos}\label{sec2}
\subsection{Rendimiento y escalabilidad}

La computación de alto rendimiento (HPC) consiste en el uso de supercomputadoras de gran capacidad para resolver problemas que requieren un alto volumen de operaciones \cite{Kurgalin2019}. En el caso del aprendizaje profundo, HPC se orienta a distribuir el entrenamiento de modelos neuronales complejos entre múltiples aceleradores\cite{Kurgalin2019}. En los últimos años, las \textit{Graphics Processing Units}(GPUs) han adquirido un papel esencial en este ámbito, gracias a su capacidad de ejecutar miles de hilos en paralelo y de procesar operaciones matriciales de forma muy eficiente. Estas propiedades son idóneas para las convoluciones y demás cálculos necesarios en redes neuronales, por lo que la infraestructura de HPC suele incluir grandes conjuntos de GPUs interconectadas mediante redes de alta velocidad\cite{hpc2023}. En este contexto, este artículo se centra en el tiempo de entrenamiento, es decir, el tiempo necesario para ajustar los parámetros de un modelo utilizando un conjunto de datos determinado\cite{MLSYS2020_411e39b1}. En este marco, definimos el Speedup, en el contexto a la comparación de sistemas, como la relación en el aumento de velocidad, que mide cuánto más rápido es un sistema en comparación con otro \cite{Molero2004} al que denominaremos de referencia. Puede definirse como la relación entre los tiempos de ejecución, y, en nuestro contexto de formación, se calcula como la relación entre el tiempo de referencia y el tiempo nuevo como se muestra en la  \textbf{Ecuación \ref{eq:equ1}}.

\begin{equation}
\text{\textit{Speedup}} = \frac{\text{Tiempo Referencia}}{\text{Tiempo Nuevo}}\label{eq:equ1}
\end{equation}
\text{Fórmula 1}\\

La Eficiencia se define como la relación entre la velocidad alcanzada y el número de aceleradores o procesadores utilizados \cite{Molero2004}. Se expresa como se muestra en la \textbf{Fórmula \ref{eq:equ2}}. Esta métrica indica la eficacia con la que se utilizan los recursos disponibles. La eficiencia alcanza un valor de 1 en casos ideales, en los que la mejora del rendimiento es proporcional al número de aceleradores empleados \cite{Molero2004}.

\begin{equation}
\text{Eficiencia} = \frac{\text{\textit{Speedup}}}{\text{Número de aceleradores}}\label{eq:equ2}
\end{equation}
\text{Fórmula 2}\\

\begin{equation}
\text{Eficiencia' }(E') = \frac{\frac{\text{Tiempo Referencia}}{\text{Tiempo Nuevo}}}{\frac{\text{Número de aceleradores}}{\text{Aceleradoras de referencia}}}\label{eq:equ3}
\end{equation}
\text{Fórmula 3}\\

La escalabilidad es la capacidad de un sistema para adaptarse a un aumento de la carga de trabajo manteniendo o mejorando su rendimiento\cite{Molero2004}. Este concepto incluye la planificación de la capacidad, que permite la asignación flexible de recursos en función de las demandas variables, como se observa en las aplicaciones distribuidas o en los sistemas bajo demanda \cite{Molero2004}.

\subsection{Clasificación}
La agrupación, es una técnica fundamental en el aprendizaje no supervisado, cuyo objetivo principal es organizar y clasificar puntos de datos de objetos o \textit{clusters} basándose en sus similitudes intrínsecas, sin necesidad de etiquetas predefinidas\cite{Xu2015}. Esta técnica se aplica ampliamente en campos como la segmentación de mercados, el análisis de datos genómicos y el procesamiento de imágenes, debido a su capacidad para descubrir patrones ocultos y estructuras subyacentes en los datos\cite{Xu2015}. Existen varios métodos de agrupación, cada uno con características y aplicaciones distintas. En este estudio, utilizamos el métodos, \textit{K-means}, el cual es uno de los algoritmos de agrupación más populares y sencillos\cite{MacQueen1965}. Divide los datos en «k» grupos predefinidos, asignando cada punto al \textit{cluster} con el centroide más cercano y minimizando iterativamente la distancia entre los puntos y su centroide asignado\cite{MacQueen1965}.

\subsection{Deep Learning }
En el campo específico del Aprendizaje Automático, los sistemas aprenden patrones directamente a partir de datos, sin ser programados explícitamente para cada tarea\cite{Xu2015}. Una de las ramas más potentes de esta área es el Aprendizaje Profundo ó \textit{Deep Learning} (DL), que se caracteriza por el uso de redes neuronales artificiales profundas con múltiples capas\cite{LeCun1998}. Estas redes han demostrado un rendimiento sobresaliente en retos como la clasificación de imágenes, la traducción automática o la generación de texto, logrando resultados que compiten e incluso superan, en muchos casos, el desempeño humano\cite{Lecun2015}. Las redes neuronales artificiales son modelos computacionales inspirados en el funcionamiento del cerebro humano, consistentes en neuronas artificiales organizadas en capas. Cada neurona recibe un conjunto de entradas, realiza una operación matemática y transmite su salida a otras neuronas \cite{Wang2003}. Estas redes pueden aprender relaciones complejas dentro de los datos ajustando las neuronas mediante algoritmos de entrenamiento\cite{Wang2003}. A diferencia de los métodos tradicionales de aprendizaje automático, en los que la extracción intervención manual \cite{LeCun1998}, DL permite la extracción automática de características de alto nivel directamente de los datos entrada\cite{Lecun2015}.

Dentro del aprendizaje profundo, se distinguen varias subáreas con objetivos y metodologías particulares, como lo son el Procesamiento de Lenguaje Natural (NLP) que se encarga de que las máquinas puedan entender, interpretar y generar lenguaje humano \cite{Devlin2018BERT}. Modelos como BERT y Llama2-LoRA se sitúan en esta categoría, aunque con matices diferenciales. BERT se especializa en tareas de comprensión del lenguaje \cite{Devlin2018BERT} y Llama2-LoRA, por su parte, apunta a la generación y manipulación de texto, y representa un modelo de lenguaje de gran escala que utiliza la técnica de Low-Rank Adaptation (LoRA), que es una técnica para entrenar eficientemente modelos grandes ajustando solo una pequeña parte de los parámetros mediante matrices de bajo rango \cite{Lermen2023LoRA70B}.

Otro campo es el de Visión por Computador, que comprende el reconocimiento y análisis automatizados de información contenida en imágenes o secuencias de vídeo. Donde el modelo RetinaNet, orientado a la detección de objetos\cite{Lin2017FocalDetection}, es un ejemplo destacado en esta rama. Se caracteriza por su arquitectura de red neuronal que combina un backbone, típicamente ResNet, el cual es un modelo para clasificación de imágenes en el campo de DL \cite{He2015,oshea2015}. Este se usa para la extracción de características con una subred de clasificación y regresión de bounding boxes, que son los marcos alrededor del objeto \cite{Lin2017FocalDetection}. 

Así mismo los Modelos Generativos enfocados en la creación de contenido sintético, tales como imágenes o secuencias de texto. Como Stable Diffusion, incluido en este documento, es un modelo generativo capaz de crear imágenes de alta calidad a partir de ruido aleatorio o de descripciones textuales\cite{Rombach2021High-ResolutionModels}.

%% file: 03-Trabajos-Relacionados.tex
\section{Trabajos Relacionados}\label{sec3}

El campo de la evaluación comparativa de hardware y software para el entrenamiento de modelos de DL ha sido objeto de numerosos estudios en los últimos años. Al hacer una segmentación y enfocarnos en los estudios que reflejan el usos comparativo en la aceleración y haciendo énfasis en el hardware, podemos encontrar trabajos como, \cite{Kang2022} en el que proponen una aproximación para la gestión eficiente de sistemas con el uso de GPU al ejecutar tareas tanto de entrenamiento como de inferencia. Su método, denominado Cost Efficient Deep Learning Job Allocation, integra técnicas de modelado agnóstico a la arquitectura y un planteamiento consciente con el costo eléctrico. El objetivo es asignar tareas heterogéneas a los nodos GPU de forma que se minimice el coste energético sin degradar el rendimiento. \cite{Kang2022}, se centra en la planificación de trabajos y la integración de métodos de right-sizing. Nuestro trabajo, en cambio, realiza un análisis comparativo de configuraciones ya definidas en MLPerf, enfocándose en la disminución de la eficiencia por GPU al crecer el número de aceleradores. \cite{Dhilleswararao2022} ofrece un amplio panorama de los aceleradores hardware para la implementación de redes neuronales profundas, comparando arquitecturas como GPU, FPGA, ASIC y CGRA. Se discuten factores de diseño, potencia, throughput, y se repasan avances en el entrenamiento y la inferencia de DNNs en distintos entornos embebidos y de alto rendimiento. En \cite{Rostam2024} los autores presentan una revisión sistemática enfocada en la optimización y aceleración de grandes modelos de lenguaje Large Language Models, (LLMs). Se discute la evolución de las bibliotecas, frameworks y estrategias de escalado para reducir los costes de entrenamiento y el tiempo de convergencia sin menoscabar el rendimiento. Si bien también se evalúa el impacto del escalado y la complejidad de modelos masivos, su énfasis es en LLMs y sus optimizaciones específicas. En la misma línea \cite{Fernandez2024} estudian la escalabilidad de modelos LLMs en configuraciones con miles de GPUs, analizando cuidadosamente estrategias de paralelismo. En nuestro caso, se incluye un modelo de lenguaje a gran escala Llama2 LoRA dentro de una comparación más amplia de cargas (BERT, Retinanet, Stable Diffusion), analizando principalmente la eficiencia al variar el número de GPUs. En \cite{Wang2019RW}, se analizan diferentes plataformas y se revelan cuellos de botella en la en la arquitectura de TPU. Por otro lado \cite{Dai2019} examina los principios de benchmarking en el contexto de dispositivos de ML y marcos de software para el entrenamiento de redes profundas. Se destacan varios enfoques de evaluación y se introducen métricas para comparar rendimiento, escalabilidad y facilidad de integración. También se hace referencia a MLPerf como organismo de referencia para medir rendimiento. En \cite{Frey2022} cuantifican cómo distintos modelos de DL escalan en GPUs con recursos de energía limitados. Ajustan modelos tipo ley de potencia para describir la forma en que el time-to-train varía según la disponibilidad de cómputo y restricciones energéticas. Sin embargo su enfoque es orientado a modelar el consumo energético. En \cite{You2017} introducen métodos algorítmicos específicos diseñados para HPC para el entrenamiento distribuido de modelos de ML que mejoran la escalabilidad. A lo largo de los años han aparecido algunas técnicas que intentan reducir el tiempo de entrenamiento, la complejidad y el consumo energético manteniendo la escalabilidad \cite{Varona2021}, a diferencia de nuestro estudio que compara los resultados de los sistemas con un benchmark específico.

%% file: 04-Method.tex
\section{Metodología}\label{sec4}

En este trabajo se analizan, de manera comparativa, los tiempos de entrenamiento reportados en MLPerf Training v4.1  \cite{LinkMLCommons,Mattson2020}  para diferentes máquinas y distintos algoritmos de aprendizaje profundo. No se llevaron a cabo experimentos directos de entrenamiento, sino que se procesaron los resultados oficiales publicados para las cargas de trabajo BERT, Llama2 LoRA, RetinaNet y Stable Diffusion. Cada sistema evaluado presenta un número variable de GPUs, lo que permite estudiar cómo se comporta el rendimiento de las configuraciones más pequeñas frente a las de escala masiva. Cada carga de trabajo presenta un conjunto de sistemas diferentes, como se muestra en la Tabla \ref{tab1}, con solo 11 máquinas presentes en las 4 cargas. 

\begin{table}[htb]
\caption{Máquinas por cargas de trabajo y sistema de referencia para cada carga}\label{tab1}
\begin{center}
{\footnotesize
\begin{tabular}{|c||p{1cm}||p{1cm}||p{1cm}|}\hline
Carga & Cantidad de Sistemas  & Cantidad GPUs referencia & Tiempo referencia  \\\hline\hline
BERT       & 28 & 2 & 366.18 \\\hline
Llama2 LoRA       & 29 & 4 & 61.87   \\\hline
RetinaNet       & 26 & 2 & 172.83  \\\hline
Stable Diffusion       & 25 & 4 & 60.52 \\\hline
\end{tabular}
}
\end{center}
\end{table}

Este trabajo se llevó a cabo utilizando la siguiente metodología. En primer lugar, se seleccionó una máquina de referencia para cada carga de trabajo caracterizada por su reducido número de GPUs y su mayor latencia en tiempo entrenamiento en comparación con el resto de equipos. A partir de esta elección, se procedió a calcular las métricas del \textit{speedup} y de E' utilizando los tiempos publicados en \cite{LinkMLCommons}, resultados públicos de MLPerf, asegurando la consistencia en la forma de medición de cada sistema y el calculo de las métricas utilizadas. En el calculó del \textit{speedup} se utilizaron los valores de la máquina de referencia para cada carga, garantizando que se pudiera cuantificar la ganancia en velocidad de entrenamiento siguiendo la Ecuación \ref{eq:equ1}. Posteriormente, se derivó al cálculo de la E', entendida como la fracción de mejora total atribuible a cada acelerador adicional y que contempla la cantidad de aceleradores de la máquina de referencia, siguiendo la Ecuación \ref{eq:equ3}. Para cada conjunto de sistemas se procedió a realizar un proceso de agrupamiento utilizando el algoritmo de K-means, con la finalidad de identificar patrones de comportamiento. Finalmente, se compararon los resultados en tablas y gráficos para identificar tendencias de escalabilidad y, sobre todo, para ubicar un rango de configuraciones óptimo en el que se obtiene un equilibrio entre la reducción del tiempo de entrenamiento y el aprovechamiento efectivo de cada GPU denotado por E'.

%% file: 05-Experimental.tex
\section{Resultados experimentales}\label{sec5}

En esta sección se describen los hallazgos principales del estudio tras comparar la eficiencia en el escalado de GPU`s en el entrenamiento de cuatro modelos diferentes de IA; BERT, Llama2, Retinanet y Stable Diffusion. Los datos muestran, de forma general, que si bien el aumento de aceleradores reduce el tiempo total de entrenamiento, la eficiencia por GPU tiende a disminuir, mostrando configuración con mayor rendimiento.

\subsection{BERT}

\begin{figure}[htb] 
\begin{center} 
  \includegraphics[width=8 cm]{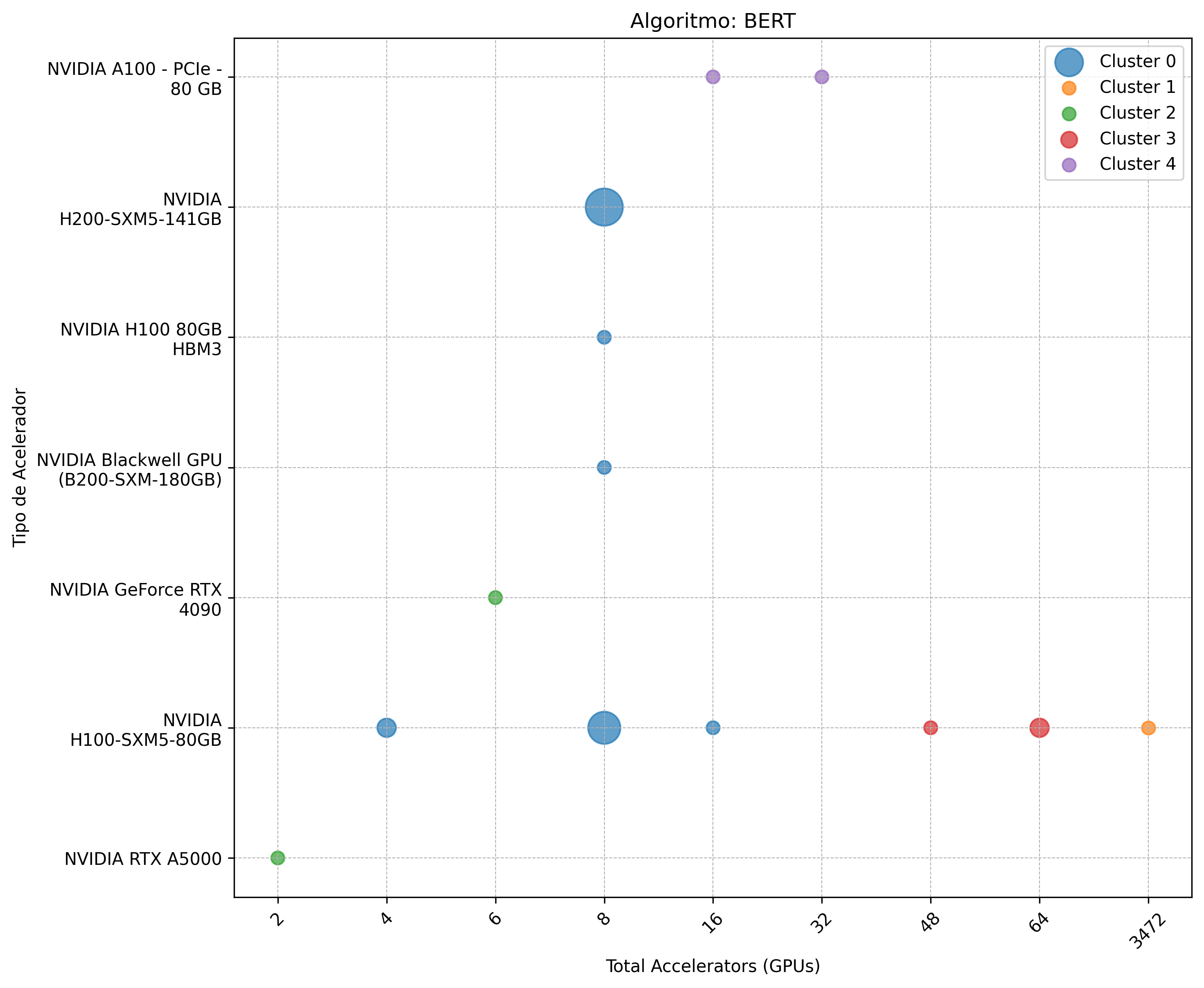}
\end{center}
\caption{Máquinas por tipo y cantidad de aceleradoras para la carga BERT \label{figA1}}
\end{figure} 

En la Fig.~\ref{figA1} se evidencia una amplia gama de configuraciones para la carga BERT, desde sistemas con tan solo 2 GPUs hasta soluciones de escalado masivo con más de 3000 aceleradores. Esto demuestra la gran heterogeneidad de sistemas utilizados para entrenar el modelo, que abarcan desde GPUs de la serie RTX hasta las más recientes familias H100 y Blackwell. El tamaño de los marcadores indica cuántas máquinas comparten cada configuración, y los colores reflejan la agrupación por clústeres. El tipo de acelerador NVIDIA H100-SXM5-80GB presenta múltiples máquinas con mayor escaldo.

\begin{figure}[htb] 
\begin{center} 
  \includegraphics[width=9 cm]{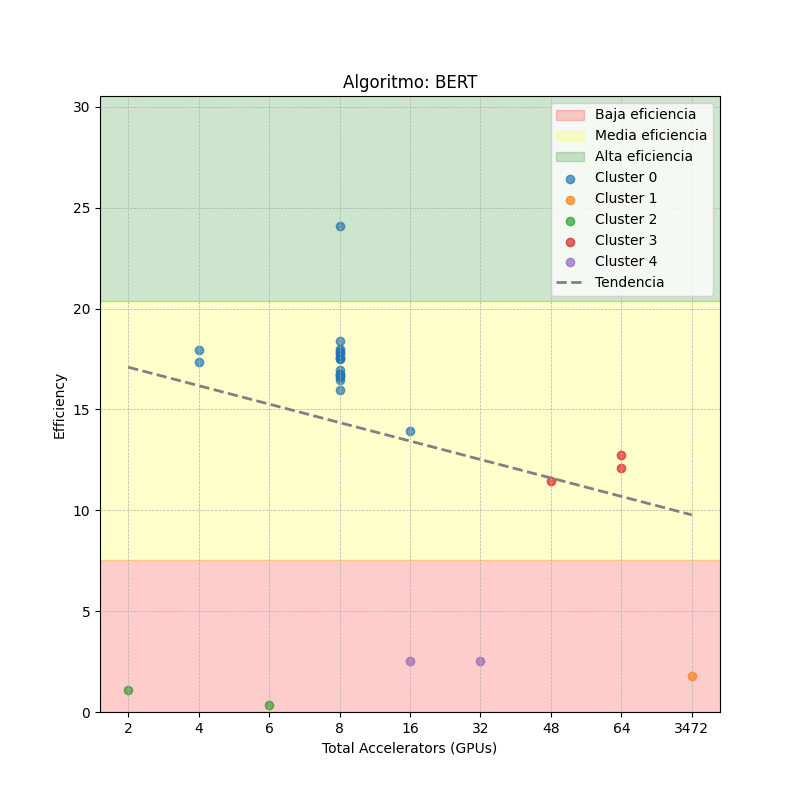}
\end{center}
\caption{Comparación de eficiencia en el algoritmo BERT \label{fig1}}
\end{figure}

Una vez calculada la eficiencia para el modelo BERT, se aprecia que las configuraciones con pocas GPUs, en torno a 8 aceleradores, exhiben los valores más altos de eficiencia, debido a que la sobrecarga de comunicación aún no resulta significativa. A partir de 16 GPUs, la eficiencia se penaliza de manera más pronunciada, tal como se ilustra en la Fig.~\ref{fig1}. No obstante, incluso en estas configuraciones más reducidas, se obtiene una mejora sustancial en el \emph{time-to-train} respecto a la máquina base, lo que sugiere que este modelo escala razonablemente bien si la prioridad es reducir el tiempo de entrenamiento. Por otro lado, se observa un hecho llamativo: en una máquina con 3472 aceleradores, la eficiencia es comparable a la de una configuración con solo 2 GPUs, lo que refleja los rendimientos decrecientes provocados posiblemente por las cargas de comunicación al llegar a escalas masivas.

\subsection{Llama2-LoRA}

\begin{figure}[htb] 
\begin{center} 
  \includegraphics[width=8 cm]{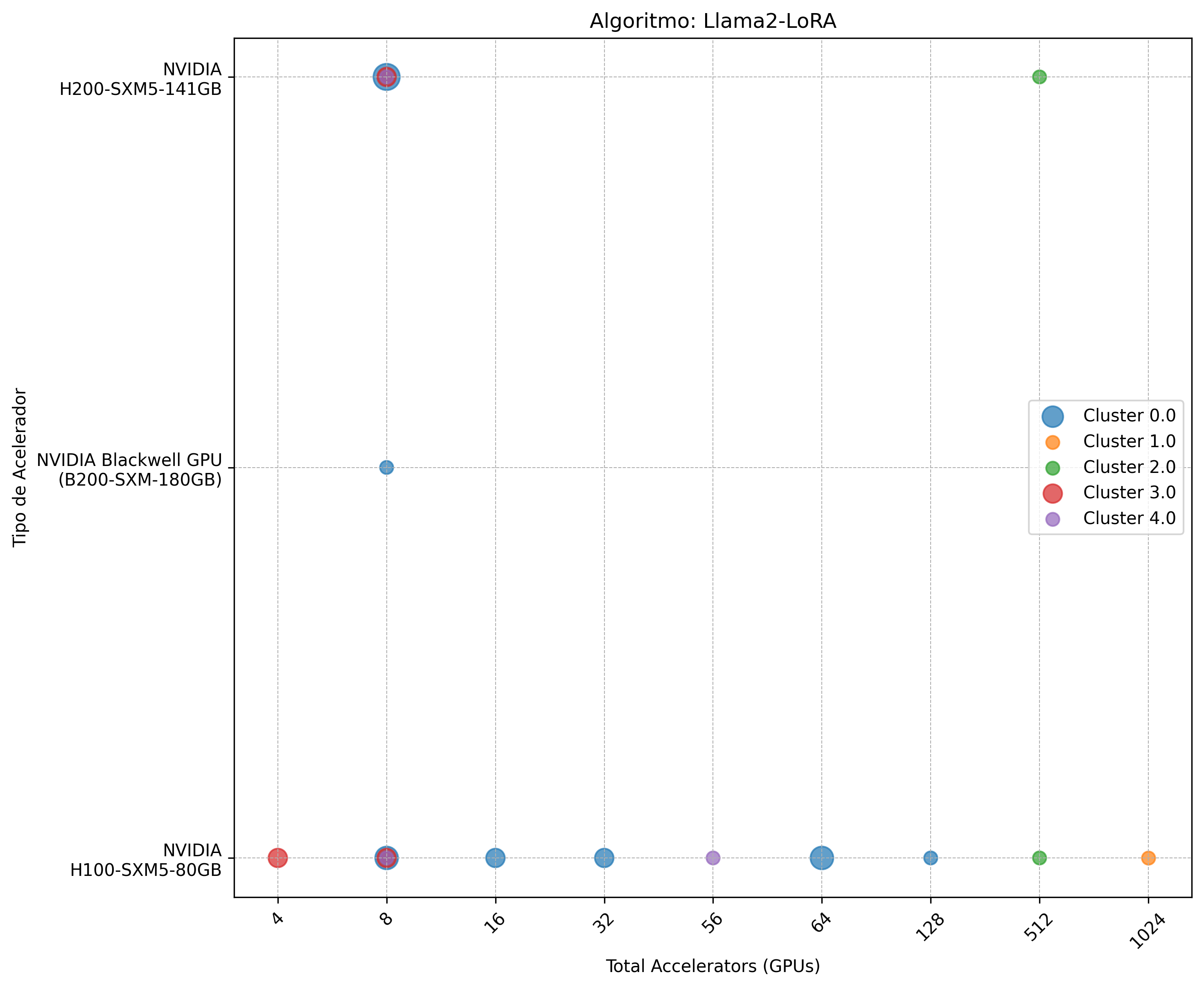}
\end{center}
\caption{Máquinas por tipo y cantidad de aceleradoras para la carga Llama2-LoRA \label{figA2}}
\end{figure} 

Para la carga de Llama2 LoRA, el número de GPUs va aproximadamente de 4 hasta 1024, como se puede ver en la Fig.~\ref{figA2}, reflejando una amplia variedad en la escala de los sistemas analizados. La mayoría de configuraciones se agrupan en torno a un rango bajo-medio de aceleradores, 4 a 64, aunque se aprecia un punto con 1024 GPUs en la parte derecha de la gráfica. La presencia de aceleradores de la NVIDIA H100-SXM5-80GB domina buena parte de los sistemas, si bien también aparecen GPU Blackwell y sistemas de la serie H200.

\begin{figure}[htb]
\begin{center}
\includegraphics[width=9 cm]{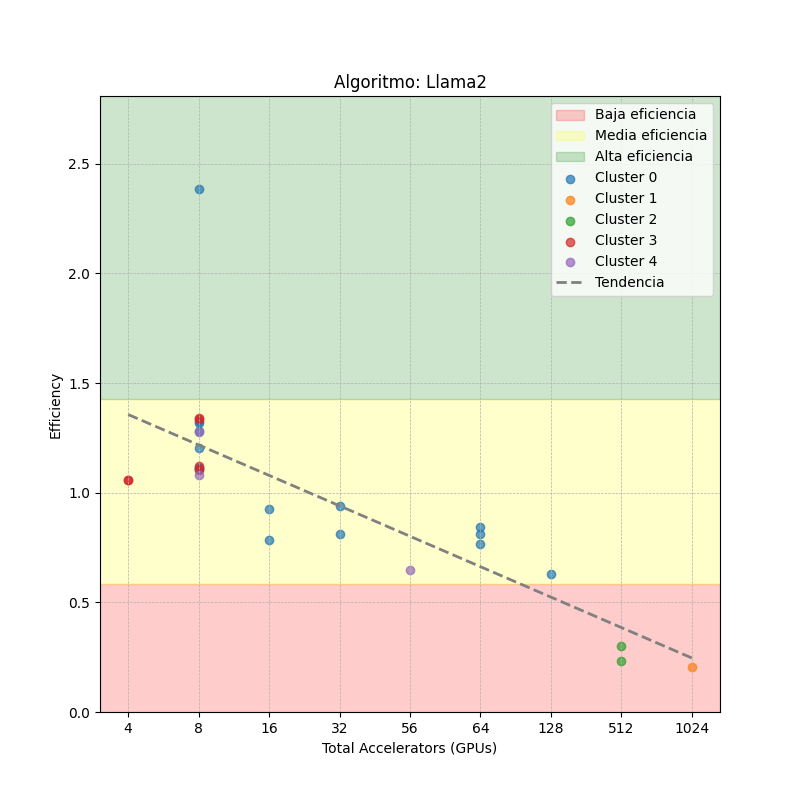}
\end{center}
\caption{Comparación de eficiencia en el algoritmo Llama2 LoRA\label{fig2}}
\end{figure} 

El comportamiento de la eficiencia en el modelo de Llama2 LoRA que exige una alta demanda de memoria y comunicaciones, se logra evidenciar una caída de eficiencia acelerada al escalar en número de GPUs. Nuevamente, en configuraciones modestas, 4 a 8 aceleradores, se aprecia una eficiencia superior a la unidad, lo que significa un buen aprovechamiento de cada GPU adicional. Sin embargo, a partir de 16 aceleradores, la curva de eficiencia decrece visiblemente, como se aprecia en la Fig. \ref{fig2}. Aun así, al comparar el time-to-train, los sistemas con mayor escalado, 1024 GPUs, concluyen el entrenamiento con una rapidez significativamente mayor, algo valioso si la prioridad es reducir drásticamente los tiempos de convergencia. De manera similar a BERT, se identificó un rango óptimo con eficiencias medias, en el que se combinan tiempos razonables con un uso equilibrado de recursos.

\subsection{RetinaNet}

\begin{figure}[htb] 
\begin{center} 
  \includegraphics[width=8 cm]{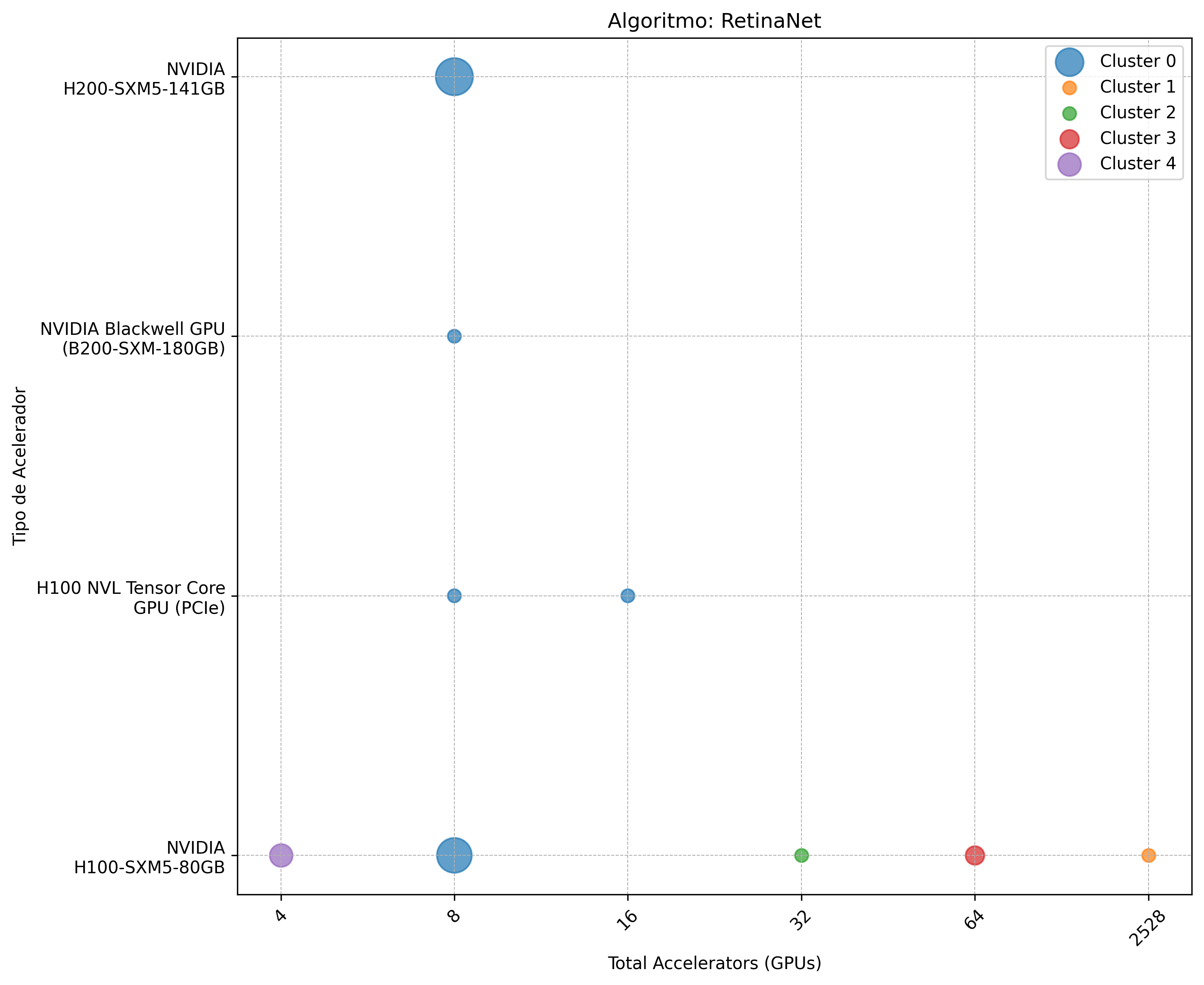}
\end{center}
\caption{Máquinas por tipo y cantidad de aceleradoras para la carga RetinaNet \label{figA3}}
\end{figure} 

En el caso de RetinaNet, la distribución de los sistemas va desde configuraciones con dos sistemas de 4 GPUs hasta otras que superan las 2000 unidades. De nuevo, se distinguen varias familias de aceleradores, H100, Blackwell, H200, lo cual revela enfoques muy diversos de escalabilidad. Los clústeres reflejan no solo la variación en la cantidad de GPUs, sino también las diferencias en los modelos de GPU empleados. Al igual que en BERT y Llama2-LoRA, el acelerador más prestente con diferente configuraciones es el NVIDIA H100-SXM5-80GB, como se ve en la Fig.~\ref{figA3}.

\begin{figure}[htb]
\begin{center}
\includegraphics[width=9 cm]{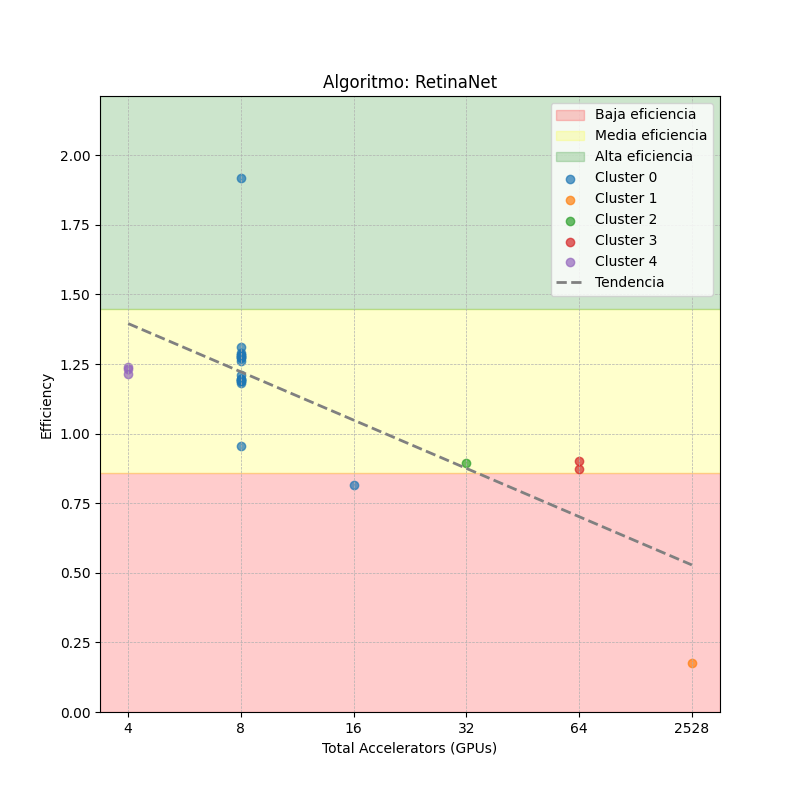}
\end{center}
\caption{Comparación de eficiencia en el algoritmo RetinaNet\label{fig3}}
\end{figure} 

Retinanet, un modelo de detección de objetos, mostró un comportamiento de escalabilidad más lineal que Llama2 LoRA, posiblemente al requerir menos sincronización de parámetros en cada iteración. No obstante, también se constató una reducción gradual en la eficiencia por GPU. Como se aprecia en la Fig. \ref{fig3}. En las máquinas de 8 GPUs, la eficiencia se mantiene en niveles medios o altos, indicando que cada GPU adicional sigue mejorando el rendimiento de forma tangible. Al sobrepasar este límite, la pendiente de la curva de eficiencia se hace más pronunciada, confirmando que los beneficios de escalar se vuelven cada vez menores. Colando en la zona de baja eficiencia a máqunas con 2528 aceleradores.

\subsection{Stable Diffusion}

\begin{figure}[htb] 
\begin{center} 
  \includegraphics[width=8 cm]{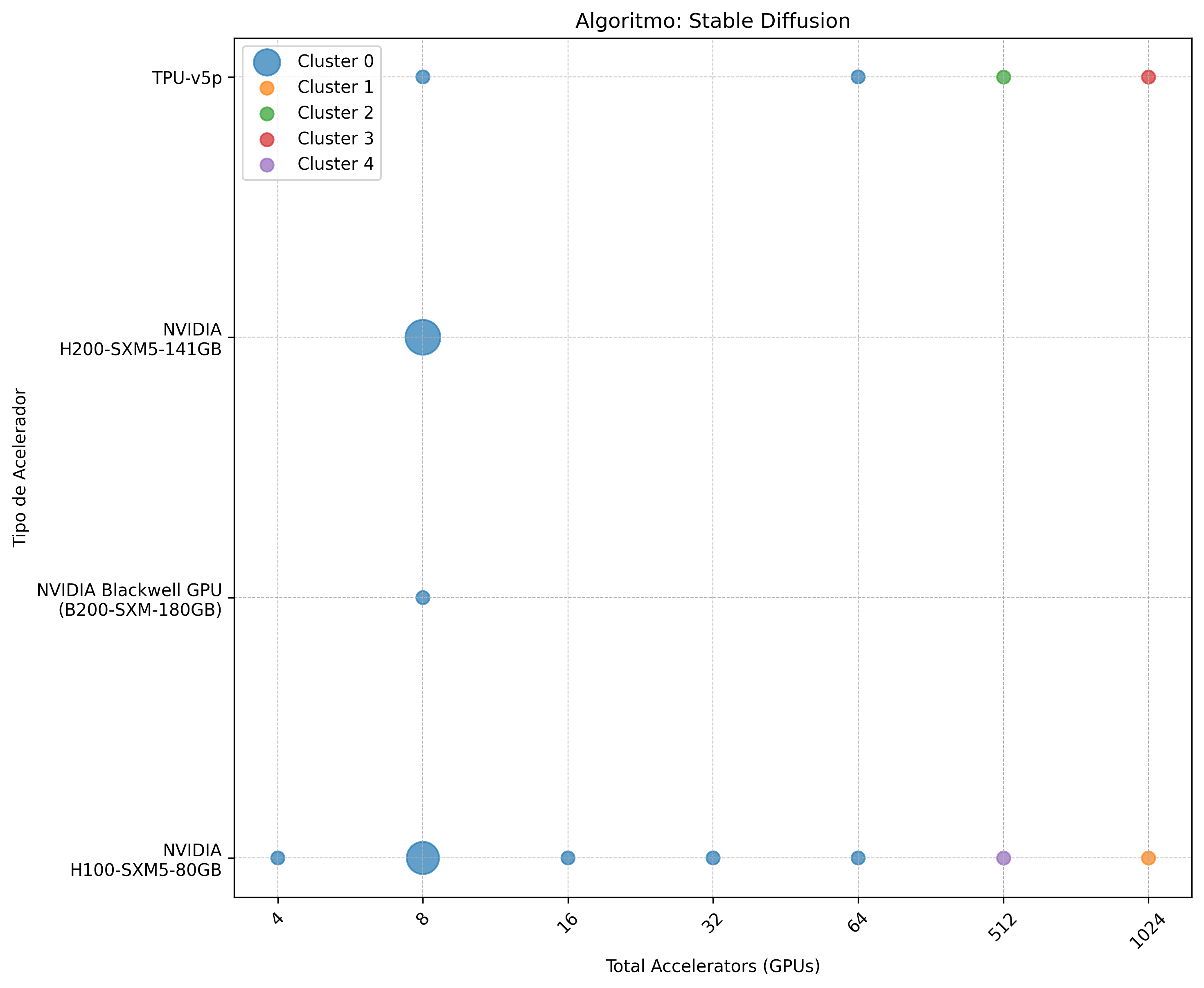}
\end{center}
\caption{Máquinas por tipo y cantidad de aceleradoras para la carga Stable Diffusion \label{figA4}}
\end{figure} 

En la Fig.~\ref{figA4}, se observan configuraciones que van de 4 a 1024 GPUs. Además de las arquitecturas de NVIDIA, la carga de Stable Diffusion, cuenta con sistemas con aceleradores TPU-v5p, lo cual añade otra dimensión de diversidad al análisis, pues se trata de un tipo diferente de acelerador. El tamaño de los puntos indica que múltiples máquinas comparten esa configuración. Nuevamente la NVIDIA H100-SXM5-80GB presenta una deversidad en escalado.

\begin{figure}[htb]
\begin{center}
\includegraphics[width=9 cm]{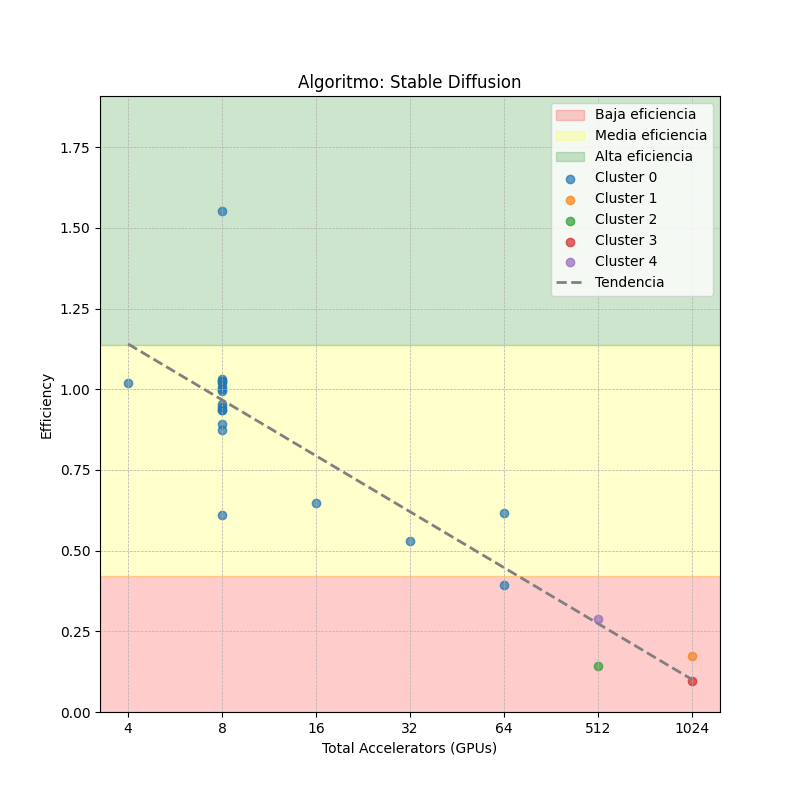}
\end{center}
\caption{Comparación de eficiencia en el algoritmo Stable Diffusion\label{fig4}}
\end{figure}

Finalmente el caso de Stable Diffusion resulta especialmente exigente, pues la combinación de un autoencoder y una U-Net con múltiples pasos de denoising incrementa la comunicación entre GPUs. Sin embargo algunas configuraciones moderadas, de 4 a 8 GPU`s presentan una eficiencia moderada y alta, posiblemente a que aprovecha la capacidad de paralelismo en el espacio latente y en la red convolucional. Sin embargo como se aprecia en la Fig. \ref{fig4}, conforme se incluyen más aceleradores, 16, 32 o incluso cientos, la eficiencia cae de manera notable, posiblemente por cuellos de botella en la comunicación de gradientes y en la sincronización de las distintas partes del modelo. Como en los demás algoritmos, se destaca un punto intermedio que maximiza la relación entre rapidez y rendimiento por GPU.

%% file: 06-Discusion.tex
\section{Discusión}\label{sec7}

Una lectura clave que se deriva del hecho de que la comunicación y sincronización de gradientes en redes de gran escala penaliza de forma notable el rendimiento individual por acelerador. Esto puede verse con especial claridad en modelos que requieren un gran intercambio de datos, como Llama2 LoRA, que maneja grandes volúmenes de parámetros y, en consecuencia, experimenta una caída acelerada de la eficiencia. Establecer un balance adecuado entre la capacidad de cómputo y la complejidad de la comunicación se puede traducir en una mejor utilización de cada GPU, lo que cobra importancia para centros de computación con restricciones de presupuesto o de consumo energético. Adicional posiblemente la distribución de las 
cargas en el entrenamiento también incide en la eficiencia. 

Estos hallazgos resaltan la necesidad de una planificación cuidadosa al dimensionar infraestructuras de cómputo. Más GPUs no siempre implican una optimización global, ya que la sobrecarga de comunicación puede neutralizar las ventajas del paralelismo. Así, la elección de la configuración “óptima” depende del objetivo principal —rapidez absoluta en el entrenamiento o maximización de la eficiencia por acelerador— y de los recursos disponibles. Sin embargo, es importante señalar que los datos utilizados provienen exclusivamente de los reportes oficiales de MLPerf Training v4.1, lo cual restringe el análisis a las configuraciones publicadas y podría representar solo una fracción del panorama real en entornos de producción.

En este sentido, el contar con un mayor número de sistemas con el mismo modelo y generación de GPU, topología de red y configuración de software, permitiría caracterizar de manera más robusta la relación entre el escalado y la eficiencia. La muestra disponible incluye equipos con distinto hardware y distintas generaciones de GPUs, lo que introduce diversidad en los resultados, pero dificulta la comparación estrictamente homogénea. Aun así, se logra describir y contrastar la ganancia de rendimiento por cada GPU adicional en la configuración establecida por MLPerf, sin profundizar en la incidencia de posibles optimizaciones de software o ajustes de hiperparámetros.

Otro aspecto relevante es que las pruebas están ceñidas a la configuración y versión del software especificados por MLPerf, sin variaciones de optimización o ajustes más finos que podrían alterar los resultados de desempeño y eficiencia. Este planteamiento, centrado principalmente en el rendimiento estandarizado, que se refiera a una visión en el que cada sistema se enfrenta a la misma prueba en las mismas condiciones. No obstante, entendemos que la seleccionan distintos optimizadores o implementación de paralelismo híbrido podría maximizar la eficiencia.

Por último, cabe destacar que la comparación no pretende evaluar el rendimiento de cada GPU de manera individual, sino describir el comportamiento global del sistema a medida que crece el número de aceleradores. Las diferencias generacionales y en capacidad de procesamiento entre las GPUs analizadas constituyen otro factor que podría incidir en la eficiencia, pero no se ha abordado en profundidad en este estudio. Pese a ello, la muestra disponible permite visualizar con claridad cómo el escalado masivo no siempre se traduce en ganancias lineales y que existen configuraciones intermedias que logran un equilibrio más adecuado entre tiempo de entrenamiento y aprovechamiento de recursos.

%% file: 07-Conclusions.tex
\section{Conclusiones}\label{sec6}

El análisis de los resultados de MLPerf Training v4.1 para diferentes máquinas y cargas de trabajo, BERT, Llama2 LoRA, Retinanet y Stable Diffusion, pone de manifiesto que en máquinas con miles de GPUs, el time-to-train es muy corto, el incremento en el número de GPUs reduce efectivamente los tiempos de entrenamiento, pero a costa de degradar la eficiencia por acelerador.

Por otro lado, aquellas configuraciones más modestas, con menos aceleradores, exhiben una eficiencia notablemente superior, si bien su tiempo global de entrenamiento aumenta. Este comportamiento apunta a la importancia de identificar puntos de equilibrio que optimicen el rendimiento sin incurrir en un sobredimensionamiento de recursos. Asimismo, se destaca que el comportamiento de la escalabilidad no es idéntico en todas las cargas, los modelos de Stable Diffusion y Llama2 LoRA, presentan una mayor caída de eficiencia con grandes volúmenes de GPUs.

En conjunto, estos hallazgos demuestran la necesidad de sopesar detenidamente tanto el time-to-train como la eficiencia por GPU al dimensionar infraestructuras de cómputo para el entrenamiento distribuido de modelos de Deep Learning. El estudio respalda la idea de que no siempre es ventajoso apuntar a configuraciones con miles de GPUs, especialmente cuando se prioriza la sostenibilidad, el control de costes o el aprovechamiento óptimo de cada acelerador. 
Como trabajo futuro, se profundizará en estrategias de paralelismo y optimizaciones de la comunicación para reducir la caída de eficiencia, así como incorporar mediciones de consumo energético y métricas de sostenibilidad que guíen la toma de decisiones de forma más holística.
\\